\documentclass{article}

\usepackage{PRIMEarxiv}

\usepackage[utf8]{inputenc} 
\usepackage[T1]{fontenc}    
\usepackage{hyperref}       
\usepackage{url}            
\usepackage{booktabs}       
\usepackage{amsfonts}       
\usepackage{nicefrac}       
\usepackage{microtype}      
\usepackage{lipsum}
\usepackage{appendix}
\usepackage[most]{tcolorbox}
\usepackage{caption}
\usepackage{tcolorbox}
\usepackage{array}
\usepackage{amsmath}
\usepackage{svg}
\usepackage{fancyhdr}       
\usepackage{graphicx}       
\graphicspath{{media/}}     

\title{A-IDE : Agent-Integrated Denoising Experts
}

\author{
  Ui Hyun Cho\thanks{These authors contributed equally to this work.},  
  Namhun Kim\footnotemark[1] \\  
  Yonsei University \\  
  Seoul, South Korea \\  
  \texttt{\{cuihyun12, ksouth0413\}@yonsei.ac.kr}  
}

\begin{document}
\maketitle

\begin{abstract}
Recent advances in deep-learning based denoising methods have improved Low-Dose CT image quality. However, due to distinct HU distributions and diverse anatomical characteristics, a single model often struggles to generalize across multiple anatomies. To address this limitation, we introduce \textbf{Agent-Integrated Denoising Experts (A-IDE)} framework, which integrates three anatomical region-specialized RED-CNN models under the management of decision-making LLM agent. The agent analyzes semantic cues from BiomedCLIP to dynamically route incoming LDCT scans to the most appropriate expert model. We highlight three major advantages of our approach. A-IDE excels in heterogeneous, data-scarce environments. The framework automatically prevents overfitting by distributing tasks among multiple experts. Finally, our LLM-driven agentic pipeline eliminates the need for manual interventions. Experimental evaluations on the Mayo-2016 dataset confirm that A-IDE achieves superior performance in RMSE, PSNR, and SSIM compared to a single unified denoiser.
\end{abstract}


\section{Introduction}
Low-Dose Computed Tomography (LDCT) imaging is crucial for reducing patient radiation exposure. However, LDCT scans suffers from increased noise and artifacts. These degradations can possibly conceal diagnostic details and impact clinical decision-making \cite{zhang2024review}. In order to ameliorate these problems, a variety of deep learning denoising approaches have shown remarkable ability to improve LDCT image quality \cite{chen2017lowdose, you2019ct, zhang2021transct, wang2021tednet}.

For instance, Residual Encoder-Decoder Convolutional Neural Network (RED-CNN) excels at capturing multi-scale image features and removing noise \cite{chen2017lowdose}. Generative Adversarial Networks (GAN) produces more realistic outputs \cite{you2019ct}. Vision transformers reconstruct high-frequency features through self-attention mechanisms \cite{zhang2021transct, wang2021tednet}. Despite these advancements, previous researches overlooked potential benefits of integrating multiple specialized denoising models into a unified framework.

A single, cumulative framework excels in denoising various anatomical regions or in data-scarce scenarios. For instance, a model trained primarily on lung CT data may not generalize well to abdominal scans. Since tissue contrasts, scanner protocols, and artifact patterns vary from structures, not a single approach excels in every situation \cite{deng2025sparse}. 

In particular, anatomical regions exhibit distinct Hounsfield Unit (HU) intensity distributions, which significantly impact denoising performance. Lung CT typically applies a windowing range of -600 to 1500 HU to emphasize air-filled structures, while abdominal CT uses a range of -160 to 240 HU to focus on on soft tissues. These variations in intensity scaling not only challenge the noise removal process \cite{Huang2020ConsideringAP, Yao2024ParallelPM}, but also underline the importance of reflecting prior anatomical labels into the denoising framework \cite{Chen2023ASCONAS, yang2025patient, Yu2024MedDiffFMAD}.

Furthermore, data scarcity and privacy issues often overfit models to train on a limited set of anatomies or characteristics due to minor unbalanced categories \cite{won2021selfsupervised,jiang2025data, saenz2023maida}. Thus, an adaptive approach that dynamically selects the most suitable denoising strategy based on anatomical context is essential. However, existing automated mechanisms lack the ability to identify and deploy region-specific denoisers effectively.

\begin{figure*}[h] 
\centering 
\includegraphics[width=\linewidth]{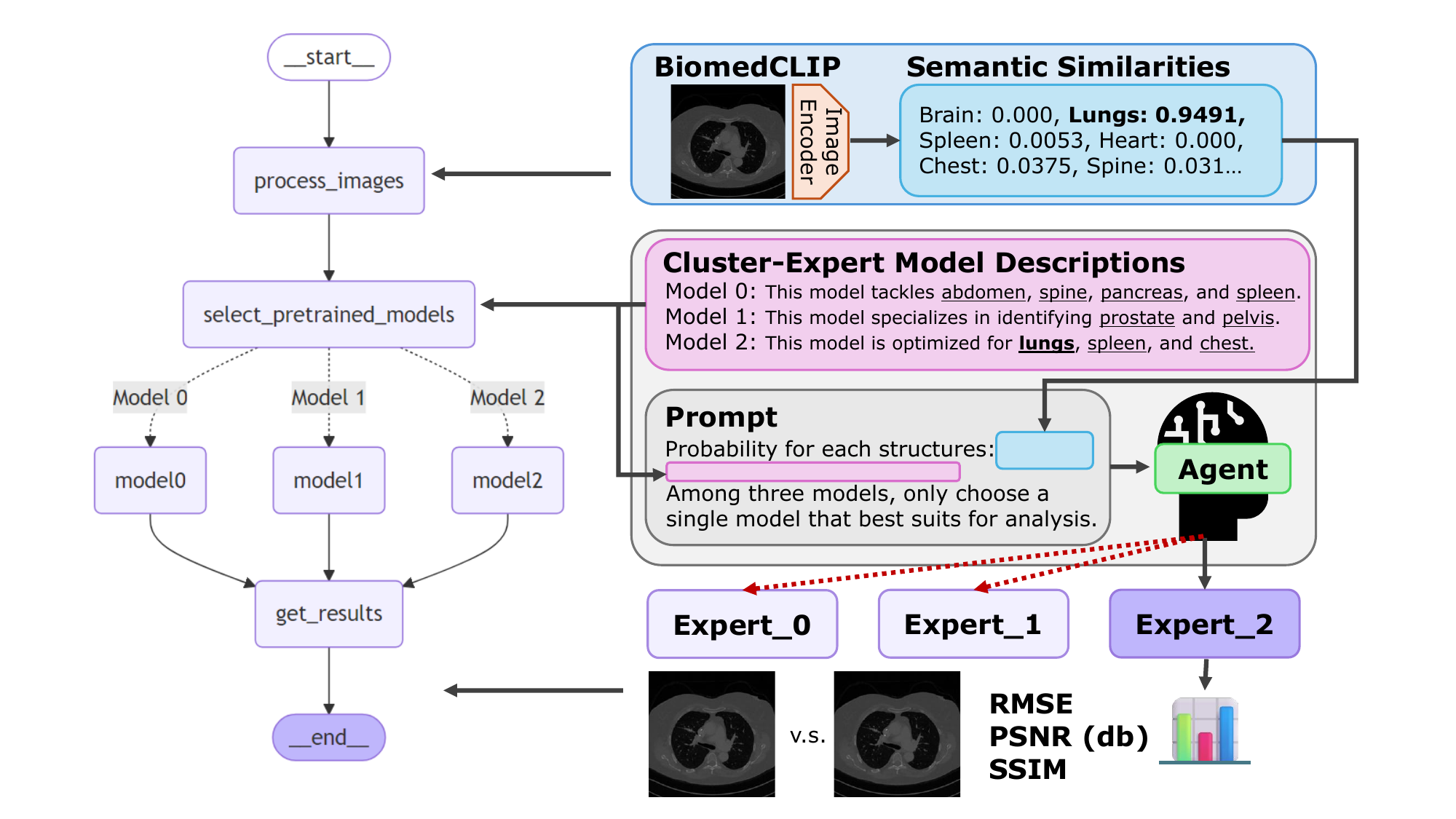} 
\caption{Overview of the \textbf{A-IDE (Agentic-Integrated Denoising Experts) Pipeline}. The input CT image is first processed by \textbf{BiomedCLIP} to generate a semantic probability distribution over anatomical structures (e.g., lungs, spleen). This distribution is then passed to a \textbf{LLM (gpt-4o) Agent} along with textual descriptions of three specialized RED-CNN denoising models: \textit{Model 0}, \textit{Model 1}, \textit{Model 2}. Based on the prompt specifying each model’s anatomical focus, the agent dynamically selects the most appropriate model for denoising. Finally, the chosen expert model reconstructs the denoised image patches and reports quantitative metrics: \textit{RMSE}, \textit{PSNR}, \textit{SSIM}. Comparative evaluation shows that automatically routing images to a cluster-specific model improves denoising performance over a single baseline model.} 
\label{fig:fig5} 
\end{figure*}

The advent of content-aware, context-specific Large Language Model (LLM) agents made intelligent interactions and automated management of sub-tasks possible \cite{shen2023hugginggpt, qin2023toolllm, masterman2024landscape}. Our \textbf{Agent-Integrated Denoising Experts (A-IDE) framework} exemplifies this trend by allocating a decision-making agent to evaluate each incoming textual context from the LDCT image and to orchestrate the optimal denoising strategy. Specifically, our agent relies on a graph-based decision making pipeline \cite{li2024mmedagent, nath2024vilam3} to get scan’s information from BiomedCLIP, and mapping it to the model that is most adept at handling the situation.

Our experimental results demonstrate that the agentic A-IDE framework, with three anatomic-specialized expert models, achieves greater Peak Signal-to-Noise Ratio (PSNR), Structural Similarity Index Measure (SSIM) performance and reduces Root Mean Square Error (RMSE) compared to the original approach. Our scores are comparable to the best single model while maintaining adaptability across diverse anatomical regions. A-IDE framework thus proves advantageous in balancing overall performance while reducing the risk that a single global model might fail on imbalanced anatomies or minor conditions.

In summary, our contribution is threefold (1) We introduce \textbf{A-IDE}, an agent-based system designed to integrate specialized LDCT denoising models into a single unified framework. (2) We employ a clustering strategy that partitions anatomical data and tailors specialized training. (3) We validate the capability of a structured and robust graphical agentic system to seamlessly orchestrate specialized models for superior denoising performance.

\section{Related Works}

\subsection{Deep-Learning Denoising Methods} 

Deep-learning methods have greatly improved LDCT denoising thanks to the strong representational capabilities neural networks provide \cite{zhang2024review}. \textbf{RED-CNN} captures local image details through multi-scale convolutions and yields strong performance with minimal structural distortion \cite{chen2017lowdose}. \textbf{GAN-CIRCLE} mitigates artifacts and maintains fine structural details by enforcing cycle-consistency via the Wasserstein distance \cite{you2019ct}. \textbf{TransCT} and \textbf{TED-NET} integrate transformer-based architectures to improve noise suppression while refining high-frequency texture details \cite{zhang2021transct, wang2021tednet}. Finally, \textbf{conditional DDPM} frameworks produce high-quality reconstructions by gradually sampling cleaner images while preserving their original structure \cite{xia2022lowdose, peng2023cbct}. However, research on integrating these architectures into a unified framework remains underexplored.

\subsection{Data Constraints and Adaptability} 

Medical denoising tasks suffer from the lack of paired LDCT and NDCT data, particularly when scanning underepresented anatomical regions \cite{won2021selfsupervised}. \textbf{Noise2Noise} attempted to use self-supervised or unsupervised approaches to mitigate the need for fully paired training data \cite{lehtinen2018noise2noise}. Despite these strategies have shown promise in natural imaging and certain medical imaging scenarios \cite{jiang2025data}, their success is mixed when the underlying anatomies differ drastically. This gap highlights the need for denoising systems to be more content-aware and adaptable.

\textit{Huang et al.} apply anatomical site labels to condition a Wasserstein Generative Adversarial Network to adapt on the specific HU distribution of each region \cite{Huang2020ConsideringAP}. \textbf{ASCON} uses a multi-scale anatomical contrastive network to capture the anatomical semantics of human tissues and to interpret the black-box denoising process \cite{Chen2023ASCONAS}. \textbf{MedDiff-FM} pretrains on a wide spectrum of anatomical data, such as head, chest, and abdomen, to improve denoising performance to specific tasks. However, these methods have yet to integrate models trained on individual anatomical regions under the management of a single, unified agent.

\subsection{Integration of LLM Agents} 

LLMs play a crucial role as a decision-making component that evaluates inputs and routes them to the most appropriate processing module. \textbf{HuggingGPT} connects ChatGPT to an array of specialized models hosted on the HuggingFace Hub \cite{shen2023hugginggpt}. It translates natural language queries into model calls and manages the whole process. \textbf{ToolLLM} extends HuggingGPT by letting LLM handle over 16,000 real-world APIs through text-based interactions \cite{qin2023toolllm}. In the medical imaging domain, \textbf{MMedAgent} and \textbf{VILA-M3} adopt LLM to select, configure, and aggregate outputs from task-specific models \cite{li2024mmedagent, nath2024vilam3}. In denoising process, \textbf{SCAN-PhysFed} framework uses LLM to generate a detailed radiology report. Then it informs a patient-level, anatomy-specific hypernetwork to perform LDCT denoising \cite{yang2025patient}. Our \textbf{A-IDE} framework takes a powerful agentic approach. It analyzes the given information and prompt through a graphical workflow, then determines the most appropriate denoising action.

\section{Experiments}

\subsection{Datasets}

Our experiment utilizes the 2016 Low-Dose X-ray CT Grand Challenge dataset from the American Association of Physicists in Medicine (Mayo-2016) \cite{mccollough2017lowdose}. Mayo-2016 dataset comprises LDCT and NDCT images from 10 anonymous patient scans. We specifically use high-resolution 1 mm slice thickness 'quarter-dose' subsets and the corresponding 'full-dose' subset images. Look up Appendix~\ref{subsec:appendix_a} for visualized examples.

\subsection{Preprocessings}

\begin{center}
    \includegraphics[width=1.0\textwidth]{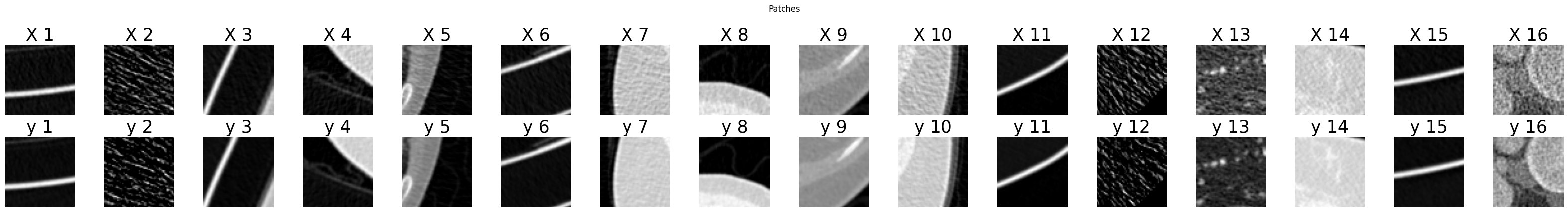}
    \captionof{figure}{Comparison of 16 quarter 1 mm and full 1 mm patches preprocessed from Mayo-2016 Dataset.}
    \label{fig:fig1}
\end{center}

We first normalize each images over the range from -1000 to 3000 Hounsfield Units (HU) using min-max scalar scaling. Pixels above or below the threshold are clipped to the corresponding maximum and minimum value. Next, we patch 512 x 512 sized images into smaller, non-overlapping 55x55 patches, generating 81 patches per image. The patchification process helps the model to handle local texture variations more effectively. Figure~\ref{fig:fig1} illustrates the patchification results, where each patched quarter-dose 1 mm image is matched to its corresponding full-dose 1 mm image.

\subsection{Clustering}

To simulate the training of anatomy-specific denoising models, we group Mayo-2016 dataset into three clusters based on captioning results from a medical vision-language embedding model, BiomedCLIP \cite{zhang2024biomedclip}. We first process each image through the model to extract embedding vectors that represent semantic similarities to twenty anatomical structures. The list of structures is shown in Figure~\ref{fig:fig2} and their visual representations can be found in Appendix~\ref{subsec:appendix_b}.

\begin{center}
    \includegraphics[width=1.0\textwidth]{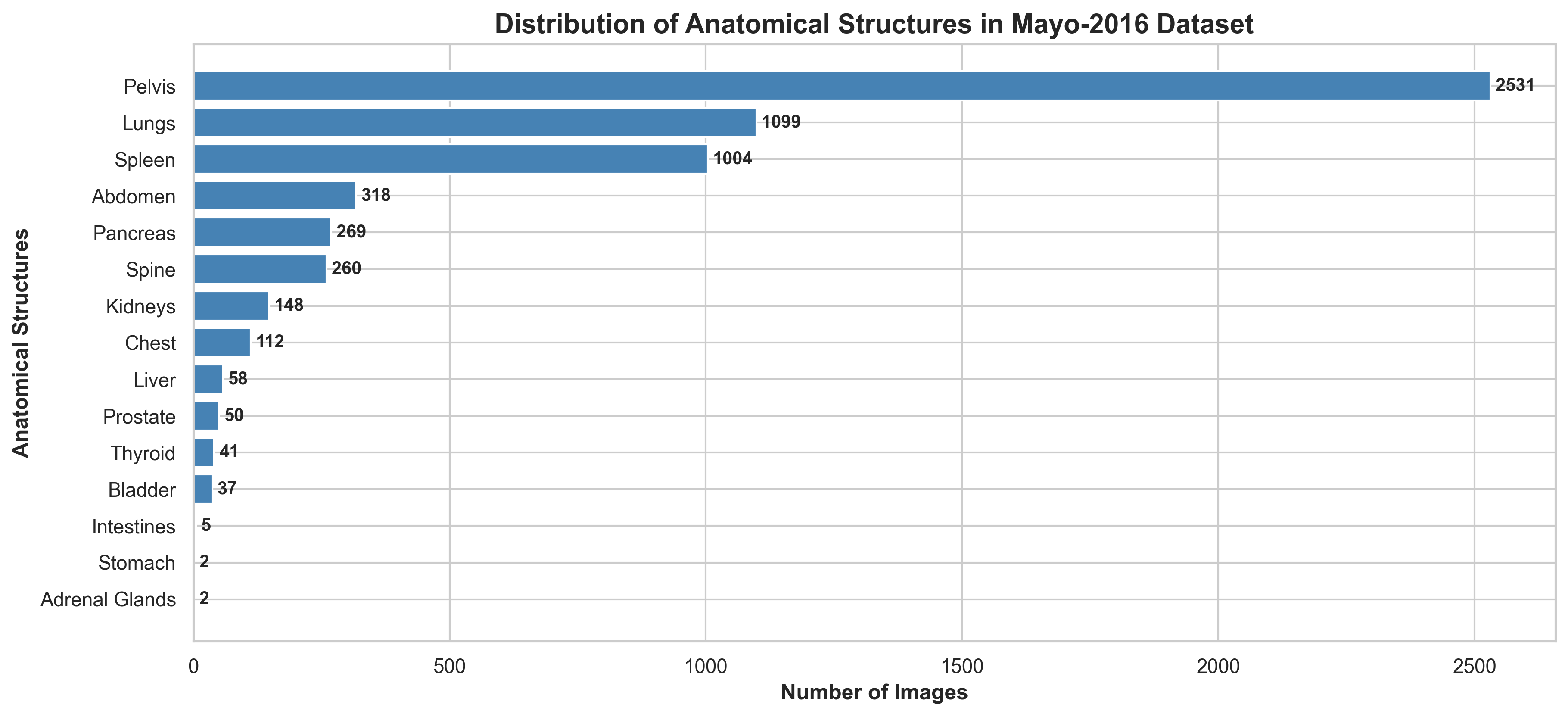}
    \captionof{figure}{Distribution of anatomical structures in the Mayo-2016 Dataset based on image counts identified by highest similarity scores from BiomedCLIP embeddings. The x-axis indicates the number of representative structures. The y-axis represents the specific anatomical types.}
    \label{fig:fig2}
\end{center}

These high-dimensional semantic similarity vectors are reduced to 2D representations using Principal Component Analysis (PCA). We then apply K-means clustering (k=3) to group these vectors based on their anatomical representations. To characterize each cluster, we randomly sample hundred images per cluster, process them with BiomedCLIP to estimate the probability of each structure, and compute the mean probability for each structure within every cluster. Figure~\ref{fig:fig3} illustrates representative examples from each cluster and the PCA results are provided in Appendix~\ref{subsec:appendix_c}.

\begin{center}
    \includegraphics[width=0.8\textwidth]{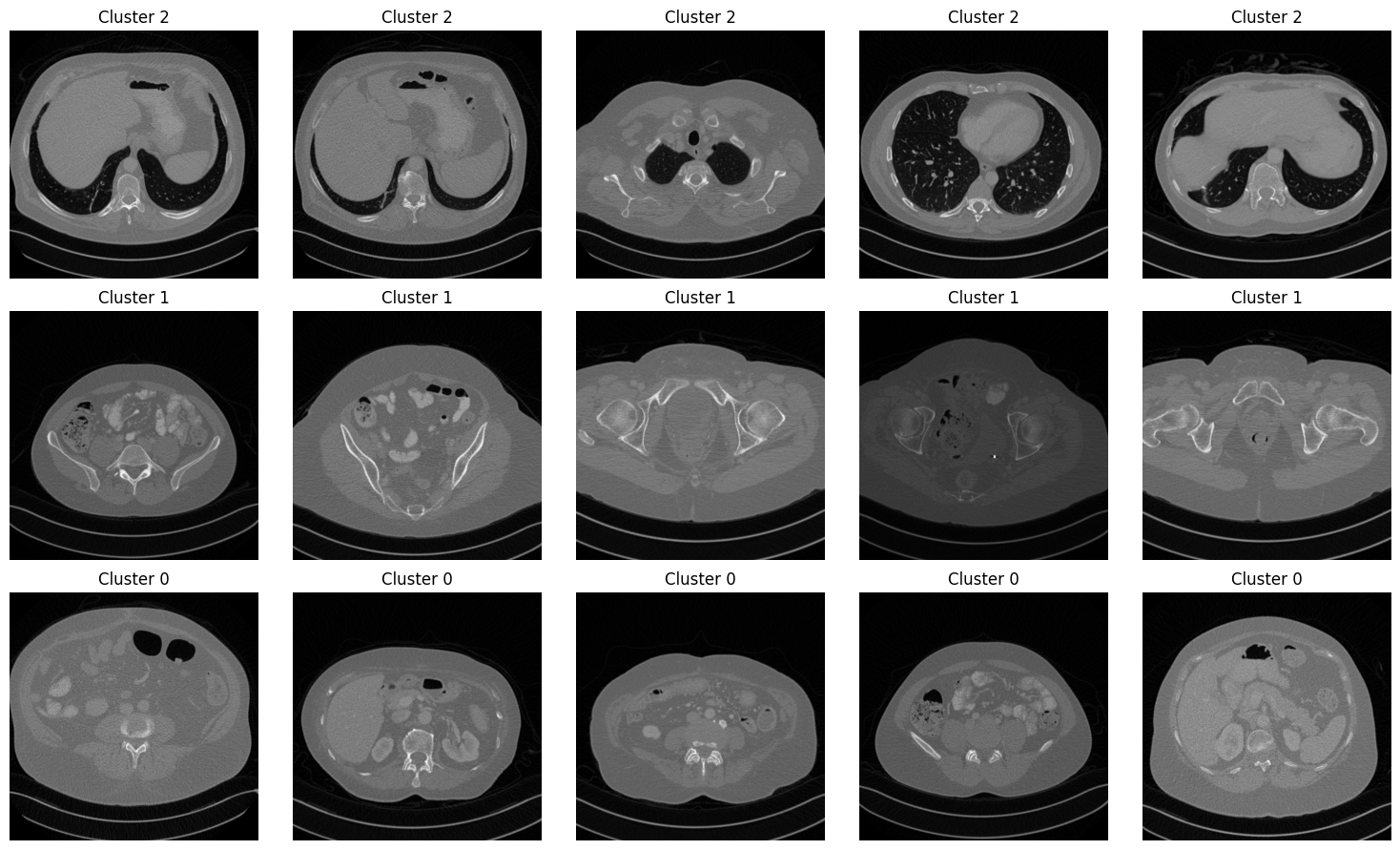}
    \captionof{figure}{Five randomly selected images per cluster. Cluster 2 focuses on lungs, cluster 1 on pelvis, and cluster 0 on abdomial organs.}
    \label{fig:fig3}
\end{center}

\begin{itemize}
\item \textbf{Cluster 0} (1,938 images) is associated with abdominal organs such as spleen, pancreas, and abdomen.
\item \textbf{Cluster 1} (2,561 images) consists mostly of pelvis.
\item \textbf{Cluster 2} (1,437 images) includes mainly lungs.
\end{itemize}

Thus, we effectively categorized LDCT images based on their anatomical content using embedding vectors from BiomedCLIP results.

\subsection{Training}

We train models under two different setups. First, we train a \textbf{baseline model} using the entire dataset. Then, we train three \textbf{cluster-expert models}, each specialized for one of the three clusters. In total, we train total four different models. We adopt the RED-CNN architecture for all training, given its effectiveness in CT image denoising. To expand the training dataset and to address data imbalance, we apply random rotation (±90°) and horizontal/vertical flips as data augmentation techniques. Additionally, for the cluster-expert models, we utilize extra rotations (30°$\sim$ 60°) to further mitigate data sparsity. Our data distribution result is following:

\begin{itemize}
\item \textbf{Baseline model} is trained on 57,579 images
\item \textbf{Cluster 0} has 37,514 images
\item \textbf{Cluster 1} has 49,603 images
\item \textbf{Cluster 2} has 27,794 images
\end{itemize}

We then split each dataset into training (64\%), validation (16\%), and test (20\%) sets. All models are trained with the Adam optimizer, an initial learning rate of 1e-5, and mean squared error as the loss function. An early stopping mechanism halts training when validation performance pleateaus below 1e-5 threshold. Furthermore,  a learning rate scheduler reduces the learning rate by half if validation loss stagnates for five epochs, down to a minimum of 1e-10. Figure~\ref{fig:fig4} illustrates loss convegence curves for baseline model and three cluster-specific models. Appendix~\ref{subsec:appendix_d} presents visual comparisons of the denoised results to the original full 1 mm images for reference.

\begin{center}
    \includegraphics[width=0.9\textwidth]{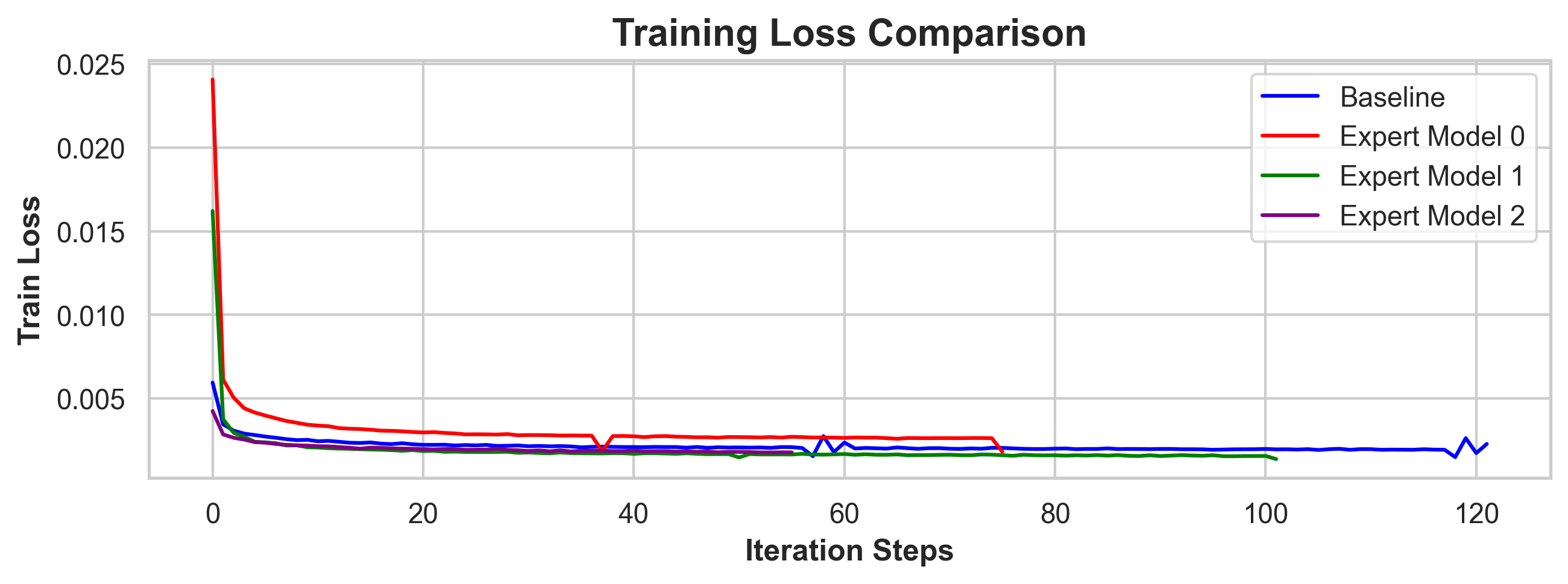}
    \captionof{figure}{Loss convergence for baseline model and three cluster-specific models. The x-axis indicates training iterations, and the y-axis represents training MSE loss.}
    \label{fig:fig4}
\end{center}

\subsection{Agents}

We set an edge-node graph system to construct an AI-driven decision making agent. This agent dynamically selects among three specialized RED-CNN denoising models based on the textual anatomical information from an individual CT image. Our agentic system is structured into three-step workflow:

\subsubsection{\texttt{process\_images}}

\begin{itemize}
    \item The input CT image undergoes feature extraction using BiomedCLIP.
    \item This node generates a semantic probability distribution over structures.
    \item The probability vectors, along with the corresponding labels, are then sent to the select\_pretrained\_models node.
\end{itemize}

\subsubsection{\texttt{select\_pretrained\_models}}

\begin{itemize}
    \item The LLM agent (backbone LLM: GPT-4o) receives the probability vectors along with textual descriptions of each specialized denoising model.
    \item Based on the prompt, the agent selects the most appropriate model. Only one among \textit{Model 0}, \textit{Model 1}, \textit{Model 2} is chosen.
    \item The agent follows the instruction from below prompt. If the agent responds \textit{Model 0}, the CT image is sent to the corresponding expert denoising model node.
\end{itemize}

\begin{tcolorbox}[
  colframe=black, colback=white, coltitle=white,
  fonttitle=\bfseries, title=System and Instruction Prompts for A-IDE,
  sharp corners,
  enhanced, width=\textwidth
]

\textbf{System Prompt:} You are \textbf{A-IDE}, an intelligent agent that chooses one of three specialized RED-CNN denoising models given their descriptions and a semantic probability distribution over anatomical structures. Reply only \textit{Model 0}, \textit{Model 1}, or \textit{Model 2}.

\vspace{0.5em}
\textbf{Prompt:} Probability for each structure:
\begin{quote}
    Brain: 0.0000, Lungs: \textbf{0.9491}, Liver: 0.0000, Stomach: 0.0000, Kidneys: 0.0000, \\ 
    Pancreas: 0.0000, Spleen: 0.0053, Heart: 0.0000, Chest: 0.0375, Abdomen: 0.0000, \\
    Pelvis: 0.0000, Spine: 0.0031, Ribs: 0.0047, Bladder: 0.0000, Prostate: 0.0000, \\
    Uterus: 0.0000, Adrenal Glands: 0.0000, Thyroid: 0.0001, Esophagus: 0.0001
\end{quote}

Among three models, only choose a single model that best suits for analysis. Be sure to choose only \textbf{one}.

\vspace{0.5em}
\textbf{Model Descriptions:}
\begin{itemize}
    \item \textit{Model 0}: This model excels at analyzing \textbf{abdomen, spine, pancreas, spleen, and kidneys}. \\
          It is fine-tuned for gastrointestinal and urological workflows.
    \item \textit{Model 1}: This model specializes in identifying \textbf{prostate and pelvis}. \\
          It offers enhanced sensitivity for small pelvic lesions and subtle urogenital abnormalities.
    \item \textit{Model 2}: This model is optimized for \textbf{lungs, spleen, and chest structure}. \\
          It is ideal for detecting respiratory conditions and lung abnormalities.
\end{itemize}

\vspace{0.5em}
\textbf{Reply only} \textit{Model 0}, \textit{Model 1}, or \textit{Model 2}

\end{tcolorbox}

\subsubsection{\texttt{get\_results}}

\begin{itemize}
    \item The selected RED-CNN expert is applied to denoise the input image patches.
    \item The final denoised image proceeds to get\_results node which prints out the reconstructed images and performs evaluation.
    \item Our A-IDE framework eliminates the need for manual intervention, making it a fully agentic, content-aware, and context-specific system that autonomously selects the optimal denoiser.
\end{itemize}

\section{Results}

\begin{table}[htbp]
\centering
\renewcommand{\arraystretch}{1.5} 
\begin{tabular}{lccc}
\hline
\textbf{Methods} & \textbf{RMSE} & \textbf{PSNR} & \textbf{SSIM}\\
\hline
Baseline & $0.097_{\scriptscriptstyle \pm 0.00164}$ & $43.06_{\scriptscriptstyle \pm 1.73}$ & $0.9557_{\scriptscriptstyle \pm 0.0125}$ \\
Expert 0 & $0.097_{\scriptscriptstyle \pm 0.00245}$ & $42.15_{\scriptscriptstyle \pm 2.28}$ & $0.9483_{\scriptscriptstyle \pm 0.0200}$ \\
Expert 1 & $\textbf{0.086}_{\scriptscriptstyle \pm 0.00165}$ & $43.33_{\scriptscriptstyle \pm 1.82}$ & $0.9576_{\scriptscriptstyle \pm 0.0107}$ \\
Expert 2 & $0.107_{\scriptscriptstyle \pm 0.00196}$ & $40.89_{\scriptscriptstyle \pm 1.39}$ & $0.9435_{\scriptscriptstyle \pm 0.0147}$ \\
\hline
A-IDE     & $0.094_{\scriptscriptstyle \pm 0.00169}$ & $\textbf{43.42}_{\scriptscriptstyle \pm 1.72}$ & $\textbf{0.9583}_{\scriptscriptstyle \pm 0.0110}$ \\
\hline
\end{tabular}
\vspace{1.0em}
\caption{Evaluation metrics of the baseline, cluster-specific experts, and the A-IDE framework (mean $\pm$ standard deviation).}
\label{tab:evaluation}
\end{table}

Table~\ref{tab:evaluation} summarizes the performance of the baseline, three cluster-specific experts, and the A-IDE framework. We use three primary metrics. RMSE quantifies the discrepancy between denoised and ground-truth images. PSNR evaluates the ratio between the maximum possible signal power and the noise. SSIM emphasizes how much the reconstructed image preserves the structural and perceptual features of the original image.

\textbf{Expert 1} achieves the lowest RMSE $0.086_{\scriptscriptstyle \pm 0.00165}$. Expert 1 model's superior performance over the baseline model highlights cluster-expert's strong capability for image denoising within its specialized anatomical domain. Furthermore, \textbf{A-IDE} outperforms all other methods in both PSNR $43.42_{\scriptscriptstyle \pm 1.72}$ and SSIM $0.9583_{\scriptscriptstyle \pm 0.0110}$. This underscores A-IDE's superior performance in generalized denoising ability. A-IDE shows consistent improvements in every metrics compared to the baseline model. This demonstrates A-IDE's robust generalization across diverse anatomical conditions.

These findings show that making intelligent agent to selectively dispatch each input to the most suitable expert model yields consistent quality gains. Our agent-based A-IDE framework uses the strengths of multiple specialized experts to balance low error rates, high signal to noise ratios, and faithful structural representation. As a result, A-IDE excels in various imaging scenarios, making it a promising solution for enhancing LDCT reconstruction in clinical scenarios.

\begin{center}
    \includegraphics[width=1.0\textwidth]{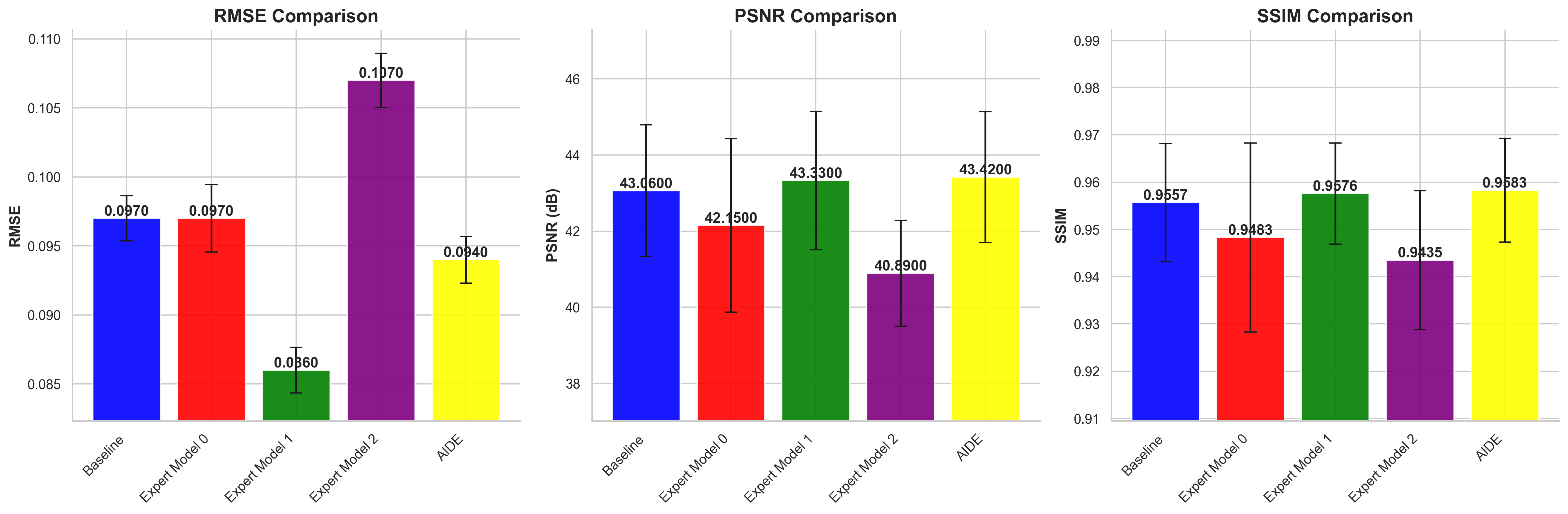}
    \captionof{figure}{Bar plots of the mean with standard deviation for the baseline model, three experts, and A-IDE. A-IDE always outforms baseline model in RMSE, PSNR, and SSIM.}
    \label{fig:fig7}
\end{center}

\section{Conclusion}

\textbf{A-IDE} framework effectively integrates multiple specialized denoising models through a decision-making LLM agent. A-IDE not only outperforms a single unified baseline in RMSE, PSNR, and SSIM metrics but also displays robust adaptability across diverse anatomical regions. The cluster-specific experts excel in their respective domains and the LLM agent coordinates these models seamlessly. Altogether, we showed that intelligent orchestration of specialized experts can achieve both precision and applicability in LDCT denoising.

Future research can explore on agent-based training paradigms. \textbf{M3Builder} dynamically controls data preprocessing, environment setup, and real-time hyperparameter tuning and reduces reliance on manual oversight in medical imaging ML workflows \cite{Feng2025M3BuilderAM}. A-IDE can incorporate multi-agent auto-tuning techniques to refine specialized cluster-experts even more effectively and further enhance LDCT denoising performance.

\bibliographystyle{unsrt}  
\bibliography{references}  

\begin{thebibliography}{10}

\bibitem{zhang2024review}
J.~Zhang, W.~Gong, L.~Ye, F.~Wang, Z.~Shangguan, and Y.~Cheng.
\newblock A review of deep learning methods for denoising of medical low-dose ct images.
\newblock {\em Computers in Biology and Medicine}, 171:108112, 2024.

\bibitem{chen2017lowdose}
Hu~Chen, Yi~Zhang, Mannudeep~K Kalra, et~al.
\newblock Low-dose ct with a residual encoder-decoder convolutional neural network.
\newblock {\em IEEE Trans. Med. Imaging}, 36(12):2524--2535, 2017.

\bibitem{you2019ct}
Chenyu You, Guang Li, Yi~Zhang, et~al.
\newblock Ct super-resolution gan constrained by the identical, residual, and cycle learning ensemble (gan-circle).
\newblock {\em IEEE Trans. Med. Imaging}, 39(1):188--203, 2019.

\bibitem{zhang2021transct}
Zhicheng Zhang, Lequan Yu, Xiaokun Liang, Wei Zhao, and Lei Xing.
\newblock Transct: Dual-path transformer for low dose computed tomography.
\newblock In {\em Medical Image Computing and Computer Assisted Intervention}, pages 55--64. Springer, 2021.

\bibitem{wang2021tednet}
D.~Wang, Z.~Wu, and H.~Yu.
\newblock Ted-net: Convolution-free t2t vision transformer-based encoder-decoder dilation network for low-dose ct denoising.
\newblock In {\em MLMI@MICCAI}, 2021.

\bibitem{deng2025sparse}
Z.~Deng and J.~Campbell.
\newblock Sparse mixture-of-experts for non-uniform noise reduction in mri images.
\newblock {\em arXiv preprint arXiv:2501.14198}, 2025.

\bibitem{Huang2020ConsideringAP}
Zhenxing Huang, Xinfeng Liu, Rongpin Wang, Jincai Chen, Ping Lu, Qiyang Zhang, Changhui Jiang, Yongfeng Yang, Xin Liu, Hairong Zheng, Dong Liang, and Zhanli Hu.
\newblock Considering anatomical prior information for low-dose ct image enhancement using attribute-augmented wasserstein generative adversarial networks.
\newblock {\em Neurocomputing}, 428:104--115, 2020.

\bibitem{Yao2024ParallelPM}
Libing Yao, Jiping Wang, Zhongyi Wu, Qiang Du, Xiaodong Yang, Ming Li, and Jian Zheng.
\newblock Parallel processing model for low-dose computed tomography image denoising.
\newblock {\em Visual Computing for Industry, Biomedicine, and Art}, 7, 2024.

\bibitem{Chen2023ASCONAS}
Zhihao Chen, Qi~Gao, Yi~Zhang, and Hongming Shan.
\newblock Ascon: Anatomy-aware supervised contrastive learning framework for low-dose ct denoising.
\newblock {\em ArXiv}, abs/2307.12225, 2023.

\bibitem{yang2025patient}
Z.~Yang, Y.~Chen, Z.~Wang, H.~Shan, Y.~Chen, and Y.~Zhang.
\newblock Patient-level anatomy meets scanning-level physics: Personalized federated low-dose ct denoising empowered by large language model.
\newblock 2025.
\newblock Unpublished manuscript.

\bibitem{Yu2024MedDiffFMAD}
Yongrui Yu, Yannian Gu, Shaoting Zhang, and Xiaofan Zhang.
\newblock Meddiff-fm: A diffusion-based foundation model for versatile medical image applications.
\newblock {\em ArXiv}, abs/2410.15432, 2024.

\bibitem{won2021selfsupervised}
D.K. Won, E.~Jung, S.~An, P.~Chikontwe, and S.H. Park.
\newblock Self-supervised learning based ct denoising using pseudo-ct image pairs.
\newblock {\em arXiv preprint arXiv:2104.02326}, 2021.

\bibitem{jiang2025data}
Y.~Jiang and V.~S.~K. Manem.
\newblock Data augmented lung cancer prediction framework using the nested case control nlst cohort.
\newblock {\em Frontiers in Oncology}, 15:1492758, 2025.

\bibitem{saenz2023maida}
A.~Saenz, E.~Chen, H.~Marklund, and P.~Rajpurkar.
\newblock The maida initiative: establishing a framework for global medical-imaging data sharing.
\newblock {\em The Lancet. Digital Health}, 2023.

\bibitem{shen2023hugginggpt}
Y.~Shen, K.~Song, X.~Tan, D.~Li, W.~Lu, and Y.T. Zhuang.
\newblock Hugginggpt: Solving ai tasks with chatgpt and its friends in hugging face.
\newblock {\em arXiv preprint arXiv:2303.17580}, 2023.

\bibitem{qin2023toolllm}
Y.~Qin, S.~Liang, Y.~Ye, K.~Zhu, L.~Yan, Y.~Lu, Y.~Lin, X.~Cong, X.~Tang, B.~Qian, S.~Zhao, R.~Tian, R.~Xie, J.~Zhou, M.H. Gerstein, D.~Li, Z.~Liu, and M.~Sun.
\newblock Toolllm: Facilitating large language models to master 16000+ real-world apis.
\newblock {\em arXiv preprint arXiv:2307.16789}, 2023.

\bibitem{masterman2024landscape}
T.~Masterman, S.~Besen, M.~Sawtell, and A.~Chao.
\newblock The landscape of emerging ai agent architectures for reasoning, planning, and tool calling: A survey.
\newblock {\em arXiv preprint arXiv:2404.11584}, 2024.

\bibitem{li2024mmedagent}
B.~Li, T.~Yan, Y.~Pan, Z.~Xu, J.~Luo, R.~Ji, S.~Liu, H.~Dong, Z.~Lin, and Y.~Wang.
\newblock Mmedagent: Learning to use medical tools with multi-modal agent.
\newblock In {\em Conference on Empirical Methods in Natural Language Processing}, 2024.

\bibitem{nath2024vilam3}
V.~Nath, W.~Li, D.~Yang, A.~Myronenko, M.~Zheng, Y.~Lu, Z.~Liu, H.~Yin, Y.M. Law, Y.~Tang, P.~Guo, C.~Zhao, Z.~Xu, Y.~He, G.~Heinrich, S.~Aylward, M.~Edgar, M.~Zephyr, P.~Molchanov, B.I. Turkbey, H.~Roth, and D.~Xu.
\newblock Vila-m3: Enhancing vision-language models with medical expert knowledge.
\newblock {\em arXiv preprint arXiv:2411.12915}, 2024.

\bibitem{xia2022lowdose}
W.~Xia, Q.~Lyu, and G.~Wang.
\newblock Low-dose ct using denoising diffusion probabilistic model for 20× speedup.
\newblock {\em arXiv preprint arXiv:2209.15136}, 2022.

\bibitem{peng2023cbct}
J.~Peng, R.L. Qiu, J.F. Wynne, C.~Chang, S.~Pan, T.~Wang, J.~Roper, T.~Liu, P.R. Patel, D.S. Yu, and X.~Yang.
\newblock Cbct-based synthetic ct image generation using conditional denoising diffusion probabilistic model.
\newblock {\em Medical Physics}, 2023.

\bibitem{lehtinen2018noise2noise}
J.~Lehtinen, J.~Munkberg, J.~Hasselgren, S.~Laine, T.~Karras, M.~Aittala, and T.~Aila.
\newblock Noise2noise: Learning image restoration without clean data.
\newblock {\em arXiv preprint arXiv:1803.04189}, 2018.

\bibitem{mccollough2017lowdose}
C.H. McCollough, A.~Bartley, R.E. Carter, B.~Chen, T.A. Drees, P.~Edwards, D.R. Holmes, A.E. Huang, F.~Khan, S.~Leng, K.~McMillan, G.J. Michalak, K.M. Nunez, L.~Yu, and J.G. Fletcher.
\newblock Low‐dose ct for the detection and classification of metastatic liver lesions: Results of the 2016 low dose ct grand challenge.
\newblock {\em Medical Physics}, 44:e339--e352, 2017.

\bibitem{zhang2024biomedclip}
Sheng Zhang, Yanbo Xu, Naoto Usuyama, Hanwen Xu, Jaspreet Bagga, Robert Tinn, Sam Preston, Rajesh Rao, Mu~Wei, Naveen Valluri, Cliff Wong, Andrea Tupini, Yu~Wang, Matt Mazzola, Swadheen Shukla, Lars Liden, Jianfeng Gao, Angela Crabtree, Brian Piening, Carlo Bifulco, Matthew~P. Lungren, Tristan Naumann, Sheng Wang, and Hoifung Poon.
\newblock A multimodal biomedical foundation model trained from fifteen million image–text pairs.
\newblock {\em NEJM AI}, 2(1), 2024.

\bibitem{Feng2025M3BuilderAM}
Jinghao Feng, Qiaoyu Zheng, Chaoyi Wu, Ziheng Zhao, Ya~Zhang, Yanfeng Wang, and Weidi Xie.
\newblock M3builder: A multi-agent system for automated machine learning in medical imaging.
\newblock 2025.

\end{thebibliography}

\appendix
\begin{samepage}
\centering
\section{Dataset Details}
\label{subsec:appendix_a}
\includegraphics[width=0.5\textwidth]{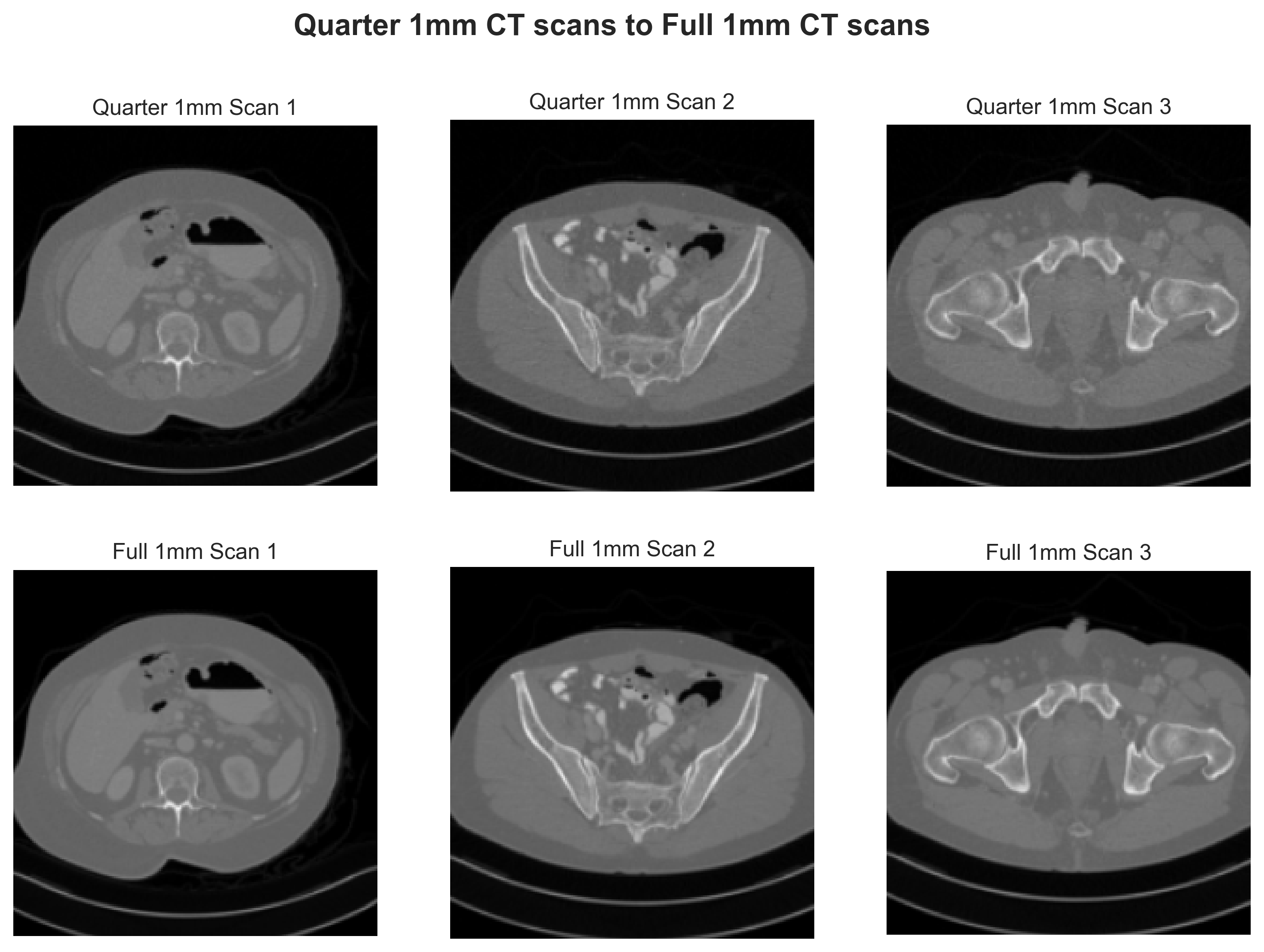}
\captionof{figure}{Examples of quarter 1 mm and full 1 mm scans of Mayo-2016 Dataset.}
\label{fig:figa}
\nopagebreak

\section{Anatomical Details}
\label{subsec:appendix_b}
\includegraphics[width=0.8\textwidth]{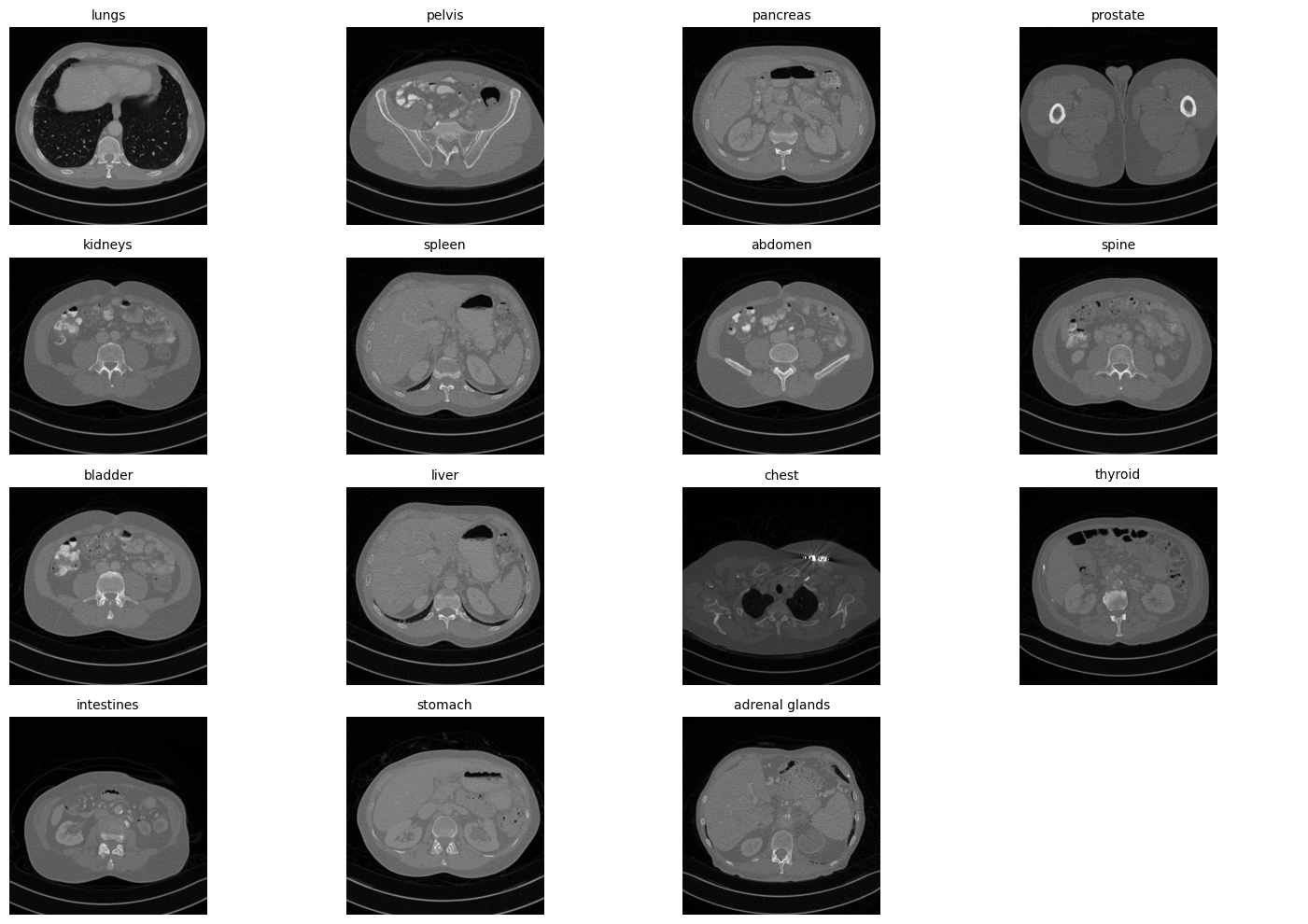}
\captionof{figure}{Examples of 15 anatomical structures in the Mayo-2016 Dataset.}
\label{fig:figb}
\nopagebreak

\section{PCA Details}
\label{subsec:appendix_c}
\includegraphics[width=0.45\textwidth]{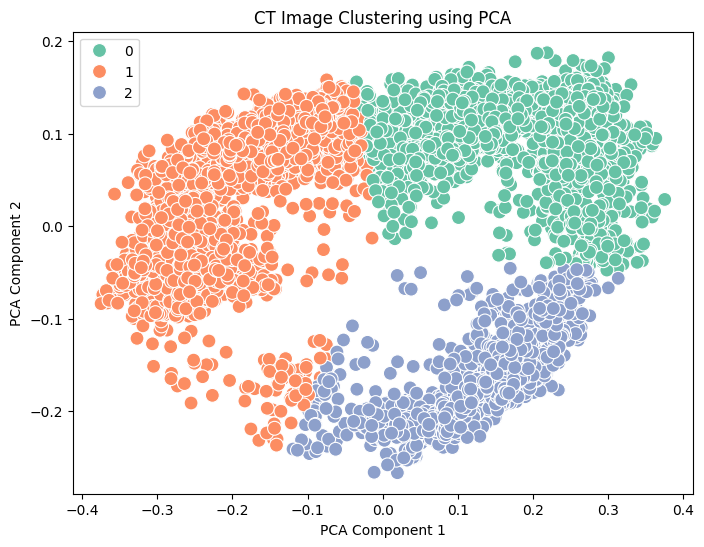}
\captionof{figure}{2D PCA projection of anatomical structure embeddings. Can identify three distinct clusters.}
\label{fig:figc}
\nopagebreak

\section{Denoising Details}
\label{subsec:appendix_d}
\includegraphics[width=0.6\textwidth]{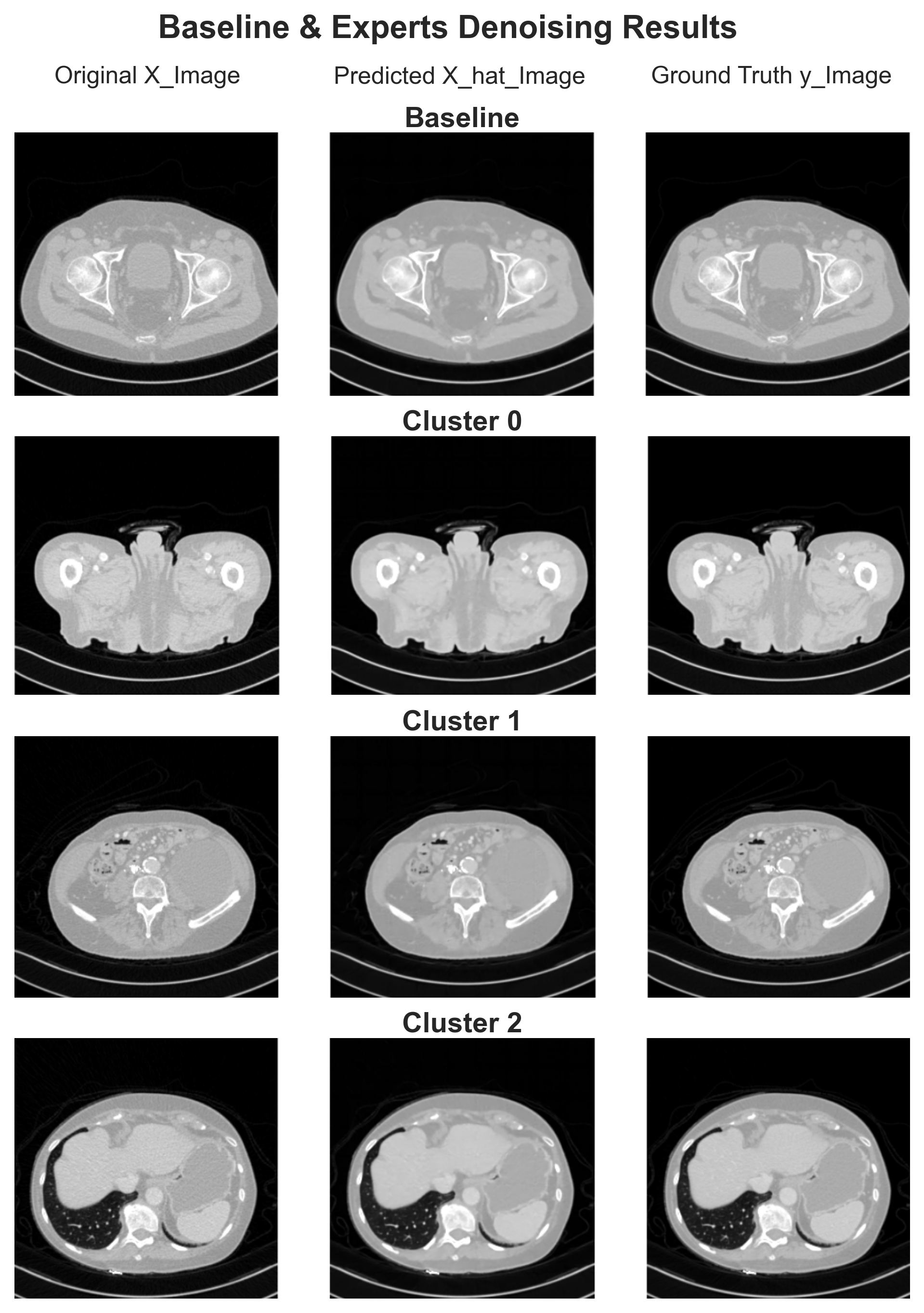}
\captionof{figure}{Visual comparisons of denoised outputs from the trained models versus the original NDCT images.}
\label{fig:figd}
\end{samepage}

\end{document}